# Dynamic Blocking and Collapsing for Gibbs Sampling

**Deepak Venugopal**
Department of Computer Science
The University of Texas at Dallas
Richardson, TX 75080, USA
dxv021000@utdallas.edu

**Vibhav Gogate**
Department of Computer Science
The University of Texas at Dallas
Richardson, TX 75080, USA
vgogate@hlt.utdallas.edu

## Abstract

In this paper, we investigate combining blocking and collapsing – two widely used strategies for improving the accuracy of Gibbs sampling – in the context of probabilistic graphical models (PGMs). We show that combining them is not straight-forward because collapsing (or eliminating variables) introduces new dependencies in the PGM and in computation-limited settings, this may adversely affect blocking. We therefore propose a principled approach for tackling this problem. Specifically, we develop two scoring functions, one each for blocking and collapsing, and formulate the problem of partitioning the variables in the PGM into blocked and collapsed subsets as simultaneously maximizing both scoring functions (i.e., a multi-objective optimization problem). We propose a dynamic, greedy algorithm for approximately solving this intractable optimization problem. Our dynamic algorithm periodically updates the partitioning into blocked and collapsed variables by leveraging correlation statistics gathered from the generated samples and enables rapid mixing by blocking together and collapsing highly correlated variables. We demonstrate experimentally the clear benefit of our dynamic approach: as more samples are drawn, our dynamic approach significantly outperforms static graph-based approaches by an order of magnitude in terms of accuracy.

## 1 Introduction

Blocking [1, 2] and collapsing [2] are two popular strategies for improving the statistical efficiency of Gibbs sampling [3] – arguably the most widely used approximate inference scheme for probabilistic graphical models (PGMs). Both these strategies trade sample quality with sample size. The hope is that the user will achieve the right balance between the two for the specific PGM at hand, improving the estimation accuracy as a result.

Unlike Gibbs sampling which samples each variable individually given others, blocked Gibbs sampling partitions the variables into disjoint groups or blocks and then *jointly samples* all variables in each block given an assignment to all other variables not in the block. Joint sampling is more expensive than sampling variables individually but the samples are of higher quality in that for a fixed sample size, the estimates based on blocked Gibbs sampling have smaller variance than the ones based on Gibbs sampling [2]. A collapsed Gibbs sampler[1] operates by *marginalizing out* a subset of variables (collapsed variables) and then generating dependent samples from the marginal distribution over the remaining variables via conventional Gibbs sampling. Marginalizing out variables is more expensive than sampling them. However, since only a sub-space is sampled, the samples are of higher quality.

Although, it is provably better to collapse a variable rather than block (group) it with other variables [2], collapsing is computationally more expensive than blocking and in practice, in many cases, the latter is feasible while the former is not. Therefore, an obvious idea is to combine blocking and collapsing, and the purpose of this paper is to investigate this combination in the context of PGMs. Specifically, the key question we seek to answer is: find a $k$-way partitioning of the variables in the PGM where each of the first $k-1$ subsets is a block and the $k$-th subset contains all the collapsed variables, such that the estimation error is minimized and the resulting algorithm is tractable. This problem is non-trivial because of the complex interplay between collapsing and blocking. For example,

**Example 1.** Consider the pair-wise Markov network (undirected PGM) given in Fig. 1(a). Let us assume that each variable in the network has $d$ values in its domain and our time and memory resource constraints dictate that we cannot incur more than $O(d^3)$ complexity. Let us further assume that we have prior knowledge that $A$, $B$, $C$, and $D$ should be blocked in order to improve the estimation accuracy (for instance, they are highly correlated or involved in deterministic constraints). Notice that we can only collapse (eliminate) $E$ and $F$ from the PGM . Otherwise, we will violate the complexity constraints. However, eliminating both $E$ and $F$ yields a clique over $A, B, C, D$ (see Fig. 1(b)) and we can no longer block these variables because the complexity of computing a joint distribution over them (using junction tree propagation) and then sampling from it is $O(d^4)$. A much better solution in this case is to collapse $F$, create two blocks $\{A, B, C, D\}$ and $\{E\}$ and perform blocked Gibbs sampling over this sub-space.

---

[1]Collapsing is often called Rao-Blackwellisation. Technically, the latter is an advanced estimator while blocking and collapsing are advanced sampling strategies. In principle, we can also use the Rao-Blackwell estimator in blocked Gibbs sampling [4]. In this paper, we will separate sampling from estimation.

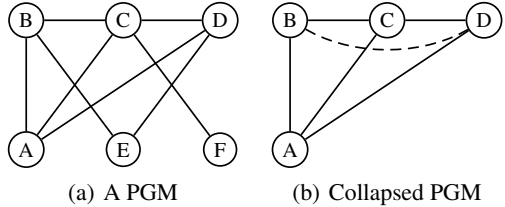

(a) A PGM  (b) Collapsed PGM

Figure 1: Example to illustrate trade-off between blocking and collapsing.

As seen from the above example, in computation-limited settings, in many cases, variables that can be blocked in the original PGM can no longer be blocked in the collapsed PGM. In other words, there is a trade-off between blocking and collapsing which needs to be taken into account while combining the two schemes. We model this tradeoff by (i) defining two integer parameters $\alpha$ and $\beta$ which bound the complexity of collapsing and blocking respectively and thus allow the user to control the number of blocked versus collapsed variables; (ii) defining two scoring functions, one each for blocking and collapsing, which favor blocks that contain variables that are highly correlated with each other and the collapsed set that contains variables which are highly correlated with other variables in the network; and (iii) casting the problem of finding the $k$-way partitioning into blocked and collapsed variables as a multi-objective optimization problem. This problem seeks to simultaneously maximize the scoring functions subject to the tractability constraints enforced by $\alpha$ and $\beta$.

The optimization problem is $\mathcal{NP}$-hard in general and therefore we propose a dynamic, greedy algorithm to solve it approximately. We integrate this algorithm with blocked-collapsed Gibbs sampling yielding a dynamic sampling algorithm. The algorithm begins by generating samples from the PGM using a feasible $k$-way partitioning computed using the (primal) graph associated with the PGM. It then periodically updates the partitioning after every $M$ samples by leveraging the correlations computed from the generated samples and performs blocked-collapsed Gibbs sampling using the new partitioning. As more samples are drawn and as the accuracy of the measured correlation increases, the underlying Markov chain is likely to mix rapidly because highly correlated variables will be either blocked together or collapsed out.

We experimentally evaluate the efficacy of our new dynamic approach on several benchmark PGMs from literature. For comparison, we use (naive) Gibbs sampling, static blocked Gibbs sampling and static blocked collapsed Gibbs sampling. Our results show that on most of the benchmark PGMs, our dynamic approach is superior to the static graph-based blocked collapsed Gibbs sampling approaches.

The rest of the paper is organized as follows. In the next section, we present background. In section 3, we define the scoring functions and our optimization problem formulation. In section 4, we present a greedy approach to solve the optimization problem and describe our dynamic Gibbs sampling algorithm. In section 5, we present related work. Section 6 describes our experimental results and we conclude in section 7.

## 2 Background

In this section, we present our notation and provide a brief overview of PGMs, Gibbs sampling, blocking, collapsing and various estimation techniques. For details, see [5, 6, 7].

A (discrete) PGM or a Markov network, denoted by $\mathcal{M}$, is a pair $\langle \mathbf{X}, \Phi \rangle$ where $\mathbf{X} = \{X_1, \ldots, X_n\}$ is a set of discrete variables (i.e., they take values from a finite domain) and $\Phi = \{\phi_1, \ldots, \phi_m\}$ is a set of positive real-valued functions (or potentials). $\mathcal{M}$ represents the probability distribution $P(\overline{\mathbf{x}}) = \frac{1}{Z} \prod_{\phi \in \Phi} \phi(\overline{\mathbf{x}}_{S(\phi)})$ where $\overline{\mathbf{x}}$ is an assignment of values to all variables in $\mathbf{X}$, $\overline{\mathbf{x}}_{S(\phi)}$ is the projection of $\overline{\mathbf{x}}$ on the scope $S(\phi)$ of $\phi$, and $Z$ is a normalization constant called the partition function. We will often abuse notation and write $\phi(\overline{\mathbf{x}}_{S(\phi)})$ as $\phi(\overline{\mathbf{x}})$. The key inference tasks in PGMs are (i) computing the partition function; (ii) computing the 1-variable marginal probabilities, i.e., computing $P(\overline{x})$ where $\overline{x}$ is an assignment of a value in the domain of $X \in \mathbf{X}$ to $X$; and (iii) computing the most probable assignment, i.e., computing $\arg\max_{\overline{\mathbf{x}}} P(\overline{\mathbf{x}})$. In this paper, we focus on the task of computing 1-variable marginals.

The primal (or interaction) graph associated with $\mathcal{M} = \langle \mathbf{X}, \Phi \rangle$, denoted by $\mathcal{G}$, is an undirected graph which has variables of $\mathcal{M}$ as its vertices and an edge between any two variables that are contained in the scope of a function $\phi \in \Phi$. The primal graph is useful because several exact inference algorithms (e.g., the junction tree algorithm [8], AND/OR graph search [9], variable (bucket) elimination [10, 11], etc.) are exponential in the *treewidth* of the primal graph and thus the primal graph can be used to quantify their complexity. The treewidth of a graph $\mathcal{G}$, denoted by $tw(\mathcal{G})$, equals the *minimum width* over all possible orderings of its vertices. The width of an ordering (either partial or full) $\pi = (X_1, \ldots, X_n)$ of a graph $\mathcal{G}$, denoted by $w(\pi, \mathcal{G})$, is the maximum degree of $X_i$ in $\mathcal{G}_{i-1}$, where $\mathcal{G} = \mathcal{G}_0, \mathcal{G}_1, \ldots, \mathcal{G}_n$ is a sequence of graphs such that $\mathcal{G}_i$ is obtained from $\mathcal{G}_{i-1}$ by adding edges so as to make the neighbor set of $X_i$ in $\mathcal{G}_{i-1}$ a clique, and then removing $X_i$ from $\mathcal{G}_i$ (i.e., eliminating $X_i$ from $\mathcal{G}_{i-1}$). For example, Fig. 1(b) shows the graph obtained by eliminating $E$ and $F$ from the graph given in Fig. 1(a). The width of the partial order $(E, F)$ is 2 while the width of the total order $(E, F, A, B, C, D)$ is 3.

Computing the treewidth of a graph is a $\mathcal{NP}$-complete problem [12]. Therefore, in practice, we often employ heuristic approaches such as the *min-fill* heuristic and *min-degree* heuristic to find an upper-bound on the treewidth. Hereafter, whenever we refer to the treewidth of a graph, we implicitly assume that we have access to a close upper-bound to the treewidth.

### 2.1 Gibbs Sampling

Given a PGM $\mathcal{M} = \langle \mathbf{X}, \Phi \rangle$, Gibbs sampling [3] begins by initializing all variables randomly, denoted by $\overline{\mathbf{x}}^{(0)}$. Then, at each iteration $j$, it randomly chooses a variable $X_i \in \mathbf{X}$ and samples a value $\overline{x}_i$ for it from the conditional distribution $P(X_i | \overline{\mathbf{x}}_{-i}^{(j-1)})$, where $\overline{\mathbf{x}}_{-i}^{(j-1)}$ denotes the projection of $\overline{\mathbf{x}}^{(j-1)}$ on all variables in the PGM other than $X_i$. The new sample is $\overline{\mathbf{x}}^{(j)} = (\overline{x}_i, \overline{\mathbf{x}}_{-i}^{(j-1)})$. The computation of the conditional distribution can be simplified by observing that in a PGM, a variable $X_i$ is conditionally independent of all other variables given its neighbors (or its *Markov blanket*) denoted by $MB(X_i)$. Formally, $P(X_i | \overline{\mathbf{x}}_{-i}) = P(X_i | \overline{\mathbf{x}}_{MB(X_i)})$. The Gibbs sampling procedure just described is called random-scan Gibbs sampling in literature. Another variation is systematic-scan Gibbs sampling in which we draw samples along a particular ordering of variables. It is known that random-scan Gibbs sampling is statistically more efficient than systematic-scan Gibbs sampling (cf. [7]).

### 2.2 Blocking and Collapsing

Blocked/Blocking Gibbs sampling [1] is an advanced sampling strategy in which some variables are sampled jointly given assignments to other variables in the PGM. Let the variables of the PGM be partitioned into disjoint groups or *blocks*, denoted by $\mathbb{B} = \{\mathbf{B}_i\}_{i=1}^{k}$, where $\mathbf{B}_i \subseteq \mathbf{X}$ and $\cup_i \mathbf{B}_i = \mathbf{X}$. Then, starting with a random assignment $\overline{\mathbf{x}}^{(0)}$ to all variables in the PGM, in each iteration $j$ of blocked Gibbs sampling, we create a new sample $\overline{\mathbf{x}}^{(j)}$ by replacing the assignment to all variables in a randomly selected block $\mathbf{B}_i$ in $\overline{\mathbf{x}}^{(j-1)}$ by a new assignment that is sampled (jointly) from the distribution, $P(\mathbf{B}_i | \overline{\mathbf{x}}_{\mathbf{X} \setminus \mathbf{B}_i}^{(j-1)})$, where $\overline{\mathbf{x}}_{\mathbf{X} \setminus \mathbf{B}_i}^{(j-1)}$ is the projection of $\overline{\mathbf{x}}^{(j-1)}$ on all variables not in $\mathbf{B}_i$. We define the Markov blanket ($MB$) of a block $\mathbf{B}_i$ as all other blocks that contain at least one variable in $MB(X_i)$, where $X_i \in \mathbf{B}_i$. Similar to Gibbs sampling, an assignment to all variables in $MB(\mathbf{B}_i)$ makes $\mathbf{B}_i$ conditionally independent of all other variables. Note that blocked Gibbs sampling is feasible only when every block $\mathbf{B}_i$ is tractable given an assignment to $MB(\mathbf{B}_i)$. These tractability constraints are often imposed in practice by putting a limit on the treewidth of the primal graph projected on the block.

Collapsing is an alternative technique for improving the accuracy of Gibbs sampling. Collapsing operates by eliminating or marginalizing out a subset of variables, say $\mathbf{C}$, from the PGM $\mathcal{M}$ yielding a collapsed PGM, $\mathcal{M}_{\mathbf{X} \setminus \mathbf{C}}$. Gibbs sampling is then performed on this smaller PGM and this improves its accuracy (because only a sub-space is sampled). In practice, collapsing is feasible only if there exists an order $\pi$ of the variables in $\mathbf{C}$ such that the width of the ordering is bounded by a small constant.

### 2.3 Estimators

Given $N$ samples $\{\overline{\mathbf{x}}^{(i)}\}_{i=1}^{N}$ drawn from the distribution $P$, we can use one of the following three estimators to compute the 1-variable marginals.

1. **Histogram estimator:**

$$\widehat{P}(\overline{x}_i) = \frac{1}{N} \sum_{j=1}^{N} \mathbb{I}_{\overline{x}_i}(\overline{\mathbf{x}}^{(j)})$$

   where $\mathbb{I}_{\overline{x}_i}(\overline{\mathbf{x}}^{(j)})$ is an indicator function which equals 1 if $\overline{x}_i$ appears in $\overline{\mathbf{x}}^{(j)}$ and 0 otherwise.

2. **Mixture Estimator:**

$$\widehat{P}(\overline{x}_i) = \frac{1}{N} \sum_{j=1}^{N} P(\overline{x}_i | \overline{\mathbf{x}}_{MB(X_i)}^{(j)})$$

3. **Rao-Blackwell Estimator:** This estimator generalizes the mixture estimator and is given by

$$\widehat{P}(\overline{x}_i) = \frac{1}{N} \sum_{j=1}^{N} P(\overline{x}_i | \overline{\mathbf{x}}_{\mathbf{R}}^{(j)})$$

   where $\mathbf{R} \subseteq \mathbf{X}$.

It has been shown that the Rao-Blackwell estimator has smaller variance than the mixture estimator which in turn has smaller variance than the histogram estimator [7] and thus the Rao-Blackwell and the mixture estimators should always be preferred. However, the Rao-Blackwell estimator requires more computation since we are essentially "ignoring" the samples on certain variables (non-sampled variables). The non-sampled variables, $\mathbf{X} \setminus \mathbf{R}$ should now be marginalized out to obtain the estimate $\widehat{P}(\overline{x}_i)$. Therefore, as the set of non-sampled variables grows larger, estimation becomes more accurate but also computationally more expensive.

All the three estimators can be used with blocked as well as collapsed Gibbs sampling. To use the Rao-Blackwell estimator with blocked Gibbs sampling, we simply find the block, say $\mathbf{B}$, in which the variable resides, set $\mathbf{R}$ equal to $\mathbf{X} \setminus \mathbf{B}$ and compute $P(\overline{x}_i | \overline{\mathbf{x}}_{\mathbf{R}}^{(j)})$ by marginalizing out all variables other than $X_i$ in the block. These computations are tractable because the block is assumed to be tractable. In collapsed Gibbs sampling, we can use the Rao-Blackwell estimator to estimate the 1-variable marginals over all the collapsed variables.

# 3 Optimally Selecting Blocked and Collapsed Variables

Integrating blocking and collapsing is tricky because they interact with each other. Moreover, we cannot collapse and block indiscriminately because for our algorithm to be practical we need to ensure that both blocking and collapsing are computationally tractable. In order to capture these constraints and the complex interplay between blocking and collapsing in a principled manner, we formulate the problem of selecting the blocks and collapsed variables as an optimization problem, defined next.

**Definition 1.** Given a PGM $\mathcal{M} = \langle \mathbf{X}, \Phi \rangle$, two scoring functions $\omega$ and $\psi$ for blocking and collapsing respectively (see sec. 3.1), and integer parameters $\alpha$, and $\beta$, find a $k$-way partition of $\mathbf{X}$ denoted by $\mathbb{X} = \mathbb{B} \cup \mathbf{C}$, where $\mathbb{B} = \{\mathbf{B}_i\}_{i=1}^{k-1}$ is a set of $k-1$ blocks and $\mathbf{C}$ is the set of collapsed variables such that both $\omega(\mathbb{B})$ and $\psi(\mathbf{C})$ are maximized, subject to two tractability constraints: (i) The minimum width of $\mathbf{C}$ in the primal graph $\mathcal{G}$ is bounded by $\alpha$; and (ii) The treewidth of $\mathcal{G}_{\setminus \mathbf{C}}$ (the graph obtained by eliminating $\mathbf{C}$ from $\mathcal{G}$) projected on each block $\mathbf{B}_i$ is bounded by $\beta$, namely, $\forall \mathbf{B}_i \in \mathbb{B}, tw(\mathcal{G}_{\setminus \mathbf{C}}(\mathbf{B}_i)) \leq \beta$.

The optimization problem just presented requires maximizing two functions and is thus an instance of a multi-objective optimization problem [13, 14]. As one can imagine, this problem is much harder than typical optimization problems in machine learning which require optimizing just one objective function. In general, there may not exist a feasible solution that simultaneously optimizes each objective function. Therefore, a reasonable approach is to find a *Pareto optimal solution*, i.e., a solution which is not dominated by any other solution in the solution space. A Pareto optimal solution cannot be improved with respect to any objective without worsening another objective.

To find Pareto optimal solutions, we will use the *lexicographic method* – a well-known approach for managing the complexity of multi-objective optimization problems. In this method, the objective functions are arranged in order of importance and we solve a sequence of single objective optimization problems. Since collapsing changes the structure of the primal graph while blocking does not, it is obvious that we should first find the collapsed variables (i.e., give more importance to the objective function for collapsing) and then compute the blocks. We will use this approach. To reduce the sensitivity of the final solution to the objective-function for collapsing, we introduce a hard penalty which penalizes solutions that result in small block sizes (since the accuracy typically increases with the block size). We describe our proposed scoring (objective) functions and the hard penalty used next.

## 3.1 Scoring Functions

We wish to design scoring functions such that they improve mixing time of the underlying Markov chain. Since the exact mixing time is hard to compute analytically, we use a heuristic scoring function that uses correlations between the variables measured periodically from the generated samples. In general, collapsing variables is much more effective when the collapsed variables exhibit high correlation with other variables in the PGM. For instance, a variable $X$ that is involved in a deterministic dependency (or constraint) with another variable $Y$ (e.g., $Y = y \rightarrow X = x$) is a good candidate for collapsing; sampling such variables likely causes the Markov chain to get stuck and hinders mixing. Similarly, blocking is effective when we jointly sample variables which are tightly correlated because sampling them separately may cause the sampler to get trapped. Moreover, we also want to minimize the number of blocks or maximize the number of variables in each block because sampling a variable jointly with other variables in a block is better than or at least as good as sampling the variables individually [7]. We quantify these desirable properties using the following scoring functions:

$$\omega(\mathbb{B}) = \frac{1}{|\mathbb{B}|} \sum_{\mathbf{B}_i \in \mathbb{B}} \sum_{X_j, X_k \in \mathbf{B}_i} D(X_j, X_k) \quad (1)$$

where $D(X_i, X_j)$ is any distance measure between the joint distribution $P(X_i, X_j)$ and the product of the marginal distributions $P(X_i)P(X_j)$.

$$\psi(\mathbf{C}) = \sum_{i=1}^{p} \frac{1}{|\mathbf{X} \setminus \mathbf{C}_{i-1}|} \sum_{X \in \mathbf{X} \setminus \mathbf{C}_{i-1}} D(C_i, X) \quad (2)$$

where $(C_1, \ldots, C_p)$ is a user-defined order on variables in $\mathbf{C}$, $\mathbf{C}_i = \{C_1, \ldots, C_i\}$ and $C_0 = \emptyset$. We use the Hellinger distance, which is a symmetric measure to compute $D(X_i, X_j)$. Formally, this distance is given by:

$$D(X_i, X_j) = \frac{1}{\sqrt{2}} \sqrt{\sum_{\overline{x}_i, \overline{x}_j} \left( \sqrt{P(\overline{x}_i, \overline{x}_j)} - \sqrt{P(\overline{x}_i)P(\overline{x}_j)} \right)^2}$$

**Algorithm 1:** Greedy-Collapse

**Input**: A PGM $\mathcal{M} = \langle \mathbf{X}, \Phi \rangle$, Integers $\alpha$, and $\gamma$
**Output**: The collapsed PGM $\mathcal{M}_{\mathbf{X}/\mathbf{C}}$ obtained by eliminating $\mathbf{C}$ from $\mathcal{M}$

1  $E = 0; \mathbf{C} = \emptyset$;
2  **repeat**
        // Let $\mathcal{G}$ be the primal graph associated with $\mathcal{M}$
3      Compute the value of the heuristic evaluation function for each vertex in $\mathcal{G}$ (see Eq. (4) );
4      Select a variable $X$ with the maximum heuristic value such that the degree $deg(X, \mathcal{G}) \leq \alpha$ where $deg(X, \mathcal{G})$ is the degree of $X$ in $\mathcal{G}$ ;
        // Let $E(X, \mathcal{G})$ be the number of new edges added to $\mathcal{G}$ by forming a clique over neighbors of $X$
5      $E = E + E(X, \mathcal{G})$;
6      Eliminate $X$ from $\mathcal{M}$;
7      $\mathbf{C} = \mathbf{C} \cup \{X\}$;
8  **until** *all vertices in $\mathcal{G}$ have degree larger than $\alpha$ or $E > \gamma$*;
9  **return** $\mathcal{M}$;

$D(X_i, X_j)$ measures the statistical dependence (correlation) between variables. Higher values indicate that the variables are statistically dependent while smaller values indicate that the variables are statistically independent. Notice that in order to compute $D(C_i, C_j)$, we need to know the 1-variable and 2-variable marginals. Their exact values are clearly not available and therefore we propose to estimate them from the generated samples.

As mentioned above, since we choose the collapsed variables before constructing the blocks, we have to penalize the feasible solutions that are likely to yield small blocks. We impose this penalty by using a hard constraint. The hard constraint disallows all feasible solutions $\mathbf{C}$ such that eliminating all variables in $\mathbf{C}$ along the ordering $(C_1, \ldots, C_p)$ will add more than $\gamma$ edges to the primal graph. Thus, $\gamma$ controls the relative importance of blocking versus collapsing. When $\gamma$ is infinite or sufficiently large, the optimal solution to the objective function for collapsing is further refined to construct the blocks. On the other hand, when $\gamma$ is small, a suboptimal solution to the objective function for collapsing, which can in turn enable higher quality blocking, is refined to construct the blocks.

## 4 Dynamic Blocked-Collapsed Gibbs Sampling

Although splitting the multi-objective optimization problem into two single objective optimization problems makes it comparatively easier to handle, it turns out that the resulting single objective optimization problems are $\mathcal{NP}$-hard. For instance, the problem of computing the set of collapsed variables includes the $\mathcal{NP}$-hard problem of computing the (weighted) treewidth (cf. [12]) as a special case. We therefore solve them using greedy methods.

### 4.1 Solving the optimization problem for Collapsing

Our greedy approach for computing the collapsed variables is given in Alg. 1. The algorithm takes as input the PGM $\mathcal{M}$, two integer parameters $\alpha$ and $\gamma$ which constrain the width of the collapsed variables (tractability constraints) and the total number of edges added to the primal graph after eliminating the collapsed variables (penalty) respectively, selects the collapsed variables, and outputs a PGM obtained by eliminating the collapsed variables.

Alg. 1 heuristically selects variables one by one for collapsing until no variables can be selected because they will violate either the tractability constraints or the (penalty) constraint on the total number of edges added. For maximizing the objective function, we want to collapse as many highly correlated variables as possible. Thus, a simple greedy approach would be to select, at each iteration, the variable $X$ with the maximum correlation score $\psi(X)$ where $\psi(X)$ is given by

$$\psi(X) = \frac{1}{|\mathbf{X}|} \sum_{X_i \in \mathbf{X}} D(X, X_i) \qquad (3)$$

However, this approach is problematic because a highly correlated variable may add several edges to the primal graph, potentially increasing its treewidth. This will in turn constrain future selections and may yield solutions which are far from optimal. In other words, at each iteration, we have to balance locally maximizing the scoring function with the number of edges added in order to have a better chance of hitting the optimum or getting close to it. We therefore use the following heuristic evaluation function to evaluate the various choices:

$$\chi(X) = \psi(X) + \left( \frac{\binom{\alpha}{2} - E(X, \mathcal{G})}{\binom{\alpha}{2}} \right) \qquad (4)$$

where $\psi(X)$ is defined in Eq. (3) and $E(X, \mathcal{G})$ is the number of new edges that will be added to $\mathcal{G}$ by forming a clique over $X$. Note that since the maximum degree of any eliminated variable is bounded by $\alpha$, the maximum number of edges that can be added is bounded by $\binom{\alpha}{2}$. Therefore, the quantity in the brackets in Eq. (4) lies between $0$ and $1$ and high values for this quantity are desirable since very few edges will be added by eliminating the particular variable ($\psi(X)$ also lies between $0$ and $1$ and high values for it are desirable too).

### 4.2 Solving the optimization problem for Blocking

Alg. 2 presents the pseudo-code for our greedy approach for constructing the blocks. The algorithm takes as input a PGM $\mathcal{M}$ and an integer parameter $\beta$ which bounds the treewidth of the primal graph of $\mathcal{M}$ projected on each block, and outputs a partitioning of the variables of $\mathcal{M}$ into blocks. The algorithm begins by having $|\mathbf{X}|$ blocks, each containing just one variable. Then it greedily merges two

**Algorithm 2:** Greedy-Block

**Input**: A PGM $\mathcal{M} = \langle \mathbf{X}, \Phi \rangle$ and Integer $\beta$
**Output**: A partition of $\mathbf{X}$ denoted by $\mathbb{B}$

1 Initialize $\mathbb{B} = \{\{X\} | X \in \mathbf{X}\}$ (each block contains just one variable);
2 **repeat**
   // Let $\mathbb{B}_{i,j}$ denote the partitioning formed from $\mathbb{B}$ by merging two blocks $\mathbf{B}_i, \mathbf{B}_j$ in $\mathbb{B}$
3   Merge two blocks $\mathbf{B}_i$ and $\mathbf{B}_j$ in $\mathbb{B}$ such that:
   1. they are in the Markov blanket of each other,
   2. $tw(\mathcal{G}(\mathbf{B}_i \cup \mathbf{B}_j)) \leq \beta$
   3. there does not exist another pair $\mathbf{B}_k, \mathbf{B}_m$ in $\mathbb{B}$ which satisfies the above two constraints and $\omega(\mathbb{B}_{k,m}) > \omega(\mathbb{B}_{i,j})$
4 **until** $\forall \mathbf{B}_i, \mathbf{B}_j \in \mathbb{B}, tw(\mathcal{G}(\mathbf{B}_i \cup \mathbf{B}_j)) > \beta$;
5 **return** $\mathbb{B}$;

---

**Algorithm 3:** Dynamic Blocked-Collapsed Sampling

**Input**: A PGM $\mathcal{M} = \langle \mathbf{X}, \Phi \rangle$; integers $T$, $M$; integers $\alpha, \beta$ and $\gamma$
**Output**: An estimate of marginal probabilities for all $X \in \mathbf{X}$

1 Initialize all 1-variable $P(\overline{x}_i)$ and 2-variable marginals $P(\overline{x}_i, \overline{x}_j)$ to zero;
2 **for** $t = 1$ to $T$ **do**
3   $\mathcal{M}_{\mathbf{X}\setminus\mathbf{C}}$ = Greedy-Collapse($\mathcal{M},\alpha,\gamma$);
4   $\mathbb{B}$ = Greedy-Block($\mathcal{M}_{\mathbf{X}\setminus\mathbf{C}},\beta$);
5   Generate $M$ samples from $\mathcal{M}_{\mathbf{X}\setminus\mathbf{C}}$ using Blocked Gibbs sampling with $\mathbb{B}$ as blocks;
6   Update all 1-variable $P(\overline{x}_i)$ and 2-variable marginals $P(\overline{x}_i, \overline{x}_j)$ using the Rao-Blackwell estimator (see Eq. (5)).
**return** $P(\overline{x}_i)$ for all variable-value combinations.

---

blocks such that they will yield the maximum increase in the score $\omega(\mathbb{B})$ under the constraint that the treewidth of the merged block is bounded by $\beta$. (Note that computing the treewidth is $\mathcal{NP}$-hard [12] and therefore in our implementation we use the min-fill algorithm to compute an upper bound on it.) To guard against merging blocks which are far away from each other in the primal graph (and thus likely to be statistically independent), we merge two blocks only if they are in the Markov blanket of each other.

### 4.3 Dynamic Blocked Collapsed Gibbs sampling

Next, we describe how to use the greedy blocking and collapsing algorithms within a Gibbs sampler, yielding an advanced sampling technique. Our proposal is summarized in Alg. 3. The algorithm takes as input a PGM $\mathcal{M}$, parameters $\alpha, \beta$ and $\gamma$ for performing blocking and collapsing, and two integers $T$ and $M$ which specify the sample size and the interval at which the statistics are updated. At termination, the algorithm outputs an estimate of all 1-variable marginal probabilities.

The algorithm maintains an estimate of 1-variable and 2-variable marginals. The 2-variable marginals are used for computing the scoring functions. At each iteration, given a $k$-way partitioning of the variables into blocked and collapsed variables, denoted by $\mathbb{B}$ and $\mathbf{C}$ respectively, the algorithm generates $M$ samples via blocked Gibbs sampling over $\mathcal{M}_{\mathbf{X}\setminus\mathbf{C}}$. After every $M$ samples the algorithm updates the blocks and collapsed variables using the greedy procedures outlined in the previous two subsections. The 1-variable and 2-variable marginals are updated using the Rao-Blackwell estimator.

Next, we describe how to update the 1-variable marginals (2-variable marginals can be updated analogously). At each iteration $t$ where $t \in \{1, T\}$, let $\{\overline{\mathbf{x}}^{(i,t)}\}_{i=1}^{M}$ be the set of $M$ samples generated via Blocked Gibbs sampling and let $\widehat{P}_t(\overline{x})$ denote the estimate of $P(X = x)$ at iteration $t$. Then $\widehat{P}_t(\overline{x})$ is given by:

$$\widehat{P}_t(\overline{x}) = \frac{(t-1)\widehat{P}_{t-1}(\overline{x}) + Q_t(\overline{x})}{t} \quad (5)$$

where $Q_t(\overline{x})$ is computed as follows. If $X \in \mathbf{C}$ is a collapsed variable, then without loss of generality, let $\mathbf{B}_k$ denote the largest block in $\mathbb{B}$. Similarly, If $X$ is a blocked variable, then without loss of generality, let $\mathbf{B}_k$ denote the block in $\mathbb{B}$ in which $X$ is present. Let $\overline{\mathbf{x}}_{-k}^{(i,t)}$ denote the projection of $\overline{\mathbf{x}}^{(i,t)}$ on all variables in $\mathbb{B} \setminus \mathbf{B}_k$. Then $Q_t$ is given as follows:

$$Q_t(\overline{x}) = \frac{1}{M} \sum_{i=1}^{M} P(\overline{x} | \overline{\mathbf{x}}_{-k}^{(i,t)}) \quad (6)$$

To compute $P(\overline{x}|\overline{\mathbf{x}}_{-k}^{(i,t)})$ we have to marginalize out all variables in $\mathbf{B}_k \cup \mathbf{C} \setminus \{X\}$. Computing this is tractable because according to our assumptions marginalizing out $\mathbf{C}$ is tractable. After marginalizing out $\mathbf{C}$, marginalizing out $\mathbf{B}_k$ is tractable because its treewidth is bounded by $\beta$.

$M$ controls the rate at which the blocks and collapsed variables are updated. Ideally, it should be greater than the burn-in period. Also, although we have assumed a constant $M$, it is easy to envision setting it using a policy in which $M$ is progressively increased as $t$ increases. From the theory of adaptive MCMC [15], it is easy to show that such a policy will ensure that estimates output by Alg. 3 will converge to $P(\overline{x}_i)$ as $T$ tends to infinity.

Note that when the correlation statistics are not available, i.e., when $t = 0$, the blocked and collapsed variables are computed by consulting the primal graph of the PGM. Thus, the blocks are constructed by randomly merging variables which are in the Markov blanket of each other; ties broken randomly. Similarly, the collapsed variables are selected along a constrained min-fill ordering (constrained by $\alpha$). Thus, if we use a time bound, namely we stop sampling after the time bound has expired, and set $M$ to be sufficiently large, Alg. 3 is equivalent to a static graph-based blocked-collapsed Gibbs sampling procedure.

| Algorithm | Blocked | Collapsed | RB | Dynamic? |
|---|---|---|---|---|
| Geman & Geman [3] | N | N | N | N |
| Jensen et al. [1] | 1 | N | N | N |
| Bidyuk & Dechter [16] | N | Y | Y | N |
| Hamze & de Freitas [4] | 2 | N | Y | N |
| Paskin [17] | M | Y | Y | N |
| Our work | M | Y | Y | Y |

Figure 2: Table comparing our work with previous work. Blocking (1: uses a single block, 2: uses 2 blocks, M: uses multiple blocks, N: not blocked), collapsing (Y/N), Rao-Blackwell Estimation (RB) (Y/N) and Dynamic (N: Static,Y: Dynamic).

## 5 Related Work

A number of earlier papers have investigated blocking and collapsing in the context of PGMs. Fig. 2 summarizes some notable ones and how they are related to our work. Blocked Gibbs sampling was first proposed by Jensen et al. [1]. The key idea in their algorithm was to create a "single block" by removing variables one by one from the primal graph until the treewidth of the remaining network is bounded by a constant and then sample this block using the junction tree algorithm. Unlike Jensen et al.'s work, we allow multiple blocks, combine collapsing with blocking and use the Rao-Blackwell estimator for computing the marginals (Jensen et al. use the histogram estimator).

Our algorithm is related to the Rao-Blackwellised blocked Gibbs sampling (RBBG) algorithm proposed by Hamze and de Freitas [4]. RBBG operates by dividing the network into two tractable tree-structured blocks and then performing Rao-Blackwellised estimation in each block. Unlike our algorithm, RBBG is applicable to grid Markov networks only. Also, unlike our algorithm, RBBG does not use multiple blocks and does not update the blocks dynamically. Moreover, RBBG does not use collapsing.

Another related work is that of Bidyuk and Dechter [16] in which the authors propose a collapsed Gibbs sampling algorithm. The key idea in their work is similar to Jensen et al.: remove variables one by one until the treewidth is bounded by a constant $w$ (the removed variables form a $w$-cutset). However, unlike Jensen et al., they use the junction tree to sample the $w$-cutset variables. Formally, let $\mathbf{W}$ be the set of $w$-cutset variables and $\mathbf{V} = \mathbf{X} \setminus \mathbf{W}$ be the set of remaining variables. Then, the junction tree is used to compute the distribution $P(W_i|\mathbf{w}_{-i})$ and sample from it. Effectively, the set $\mathbf{V}$ is always collapsed out. A key drawback of this algorithm is that the junction tree algorithm must be run from scratch for sampling each $w$-cutset variable and as a result the algorithm can be quite slow. In this paper, we save time by marginalizing out a subset of variables before running the junction tree algorithm (i.e., marginalization is a pre-processing step before sampling). Also, unlike our work, the Bidyuk and Dechter algorithm does not use blocking and is not dynamic.

The sample propagation algorithm of Mark Paskin [17] is the only blocked-collapsed algorithm for PGMs that we are aware of. The algorithm integrates sampling with message passing in a junction tree. The key idea is to walk the clusters of a junction tree, sampling some of the current cluster's variables and then passing a message to one of its neighbors. The algorithm designates a subset of variables for sampling and marginalizes out the remaining variables by performing message passing over the junction. In that sense, sample propagation is similar to (but more efficient than) Bidyuk and Dechter's algorithm. The only difference is that variables within each cluster are sampled jointly (or blocked) if the cluster size is small enough or sampled using Metropolis-Hastings otherwise. Since the blocks in sample propagation are confined to the clusters of a junction tree, they can be much smaller than the blocks used in our algorithm. Also, this algorithm is not dynamic.

Our work is related to the recent work of Venugopal and Gogate [18], who cast the problem of constructing blocks in lifted Gibbs sampling as an optimization problem, but do not update the blocks dynamically. Finally, our work is related to parallel Gibbs sampling by Gonzalez et al. [19] who use likelihood estimates to compute the blocks.

## 6 Experiments

In this section, we experimentally evaluate the performance of the following algorithms on several benchmark PGMs: (*a*) Naive Gibbs sampling (`Gibbs`); (*b*) Static Blocked Gibbs sampling (`SBG`); (*c*) Static blocked collapsed Gibbs sampling (`SBCG`); and (*d*) Dynamic blocked collapsed Gibbs sampling (`DBCG`) . `SBG` is similar to the algorithm of Hamze and de Freitas [4] except that we allow multiple blocks and do not constrain the blocks to be tree structured. `SBCG` is an advanced version of Paskin's sample propagation algorithm [17]. We implemented `SBG` and `SBCG` by setting $M$ to a sufficiently large value i.e., these methods consult only the primal graph of the PGM to choose the blocks and collapsed variables. To compute marginals, we use the Rao-Blackwell estimator in `SBG`, `SBCG` and `DBCG`, and the mixture estimator in `Gibbs`.

We tested the algorithms on several benchmark PGMs used in the UAI-2008 (graph-mod.ics.uci.edu/uai08/Evaluation/Report), and the UAI-2010 (cs.huji.ac.il/project/PASCAL) probabilistic inference competitions. For each network, we measured performance using the average Hellinger distance between the true 1-variable marginals and the estimated 1-variable marginals. We performed our experiments on a centOS machine with a quad-core processor and 8GB RAM. Each algorithm was run for 500 seconds on each benchmark for the task of estimating 1-variable marginals. In `DBCG` we set $\alpha = \beta = 8$, $\gamma = 50 \times \alpha$ and $M = 1000$. We evaluate the impact of $\alpha$, $\beta$ and $\gamma$ in the next sub-section.

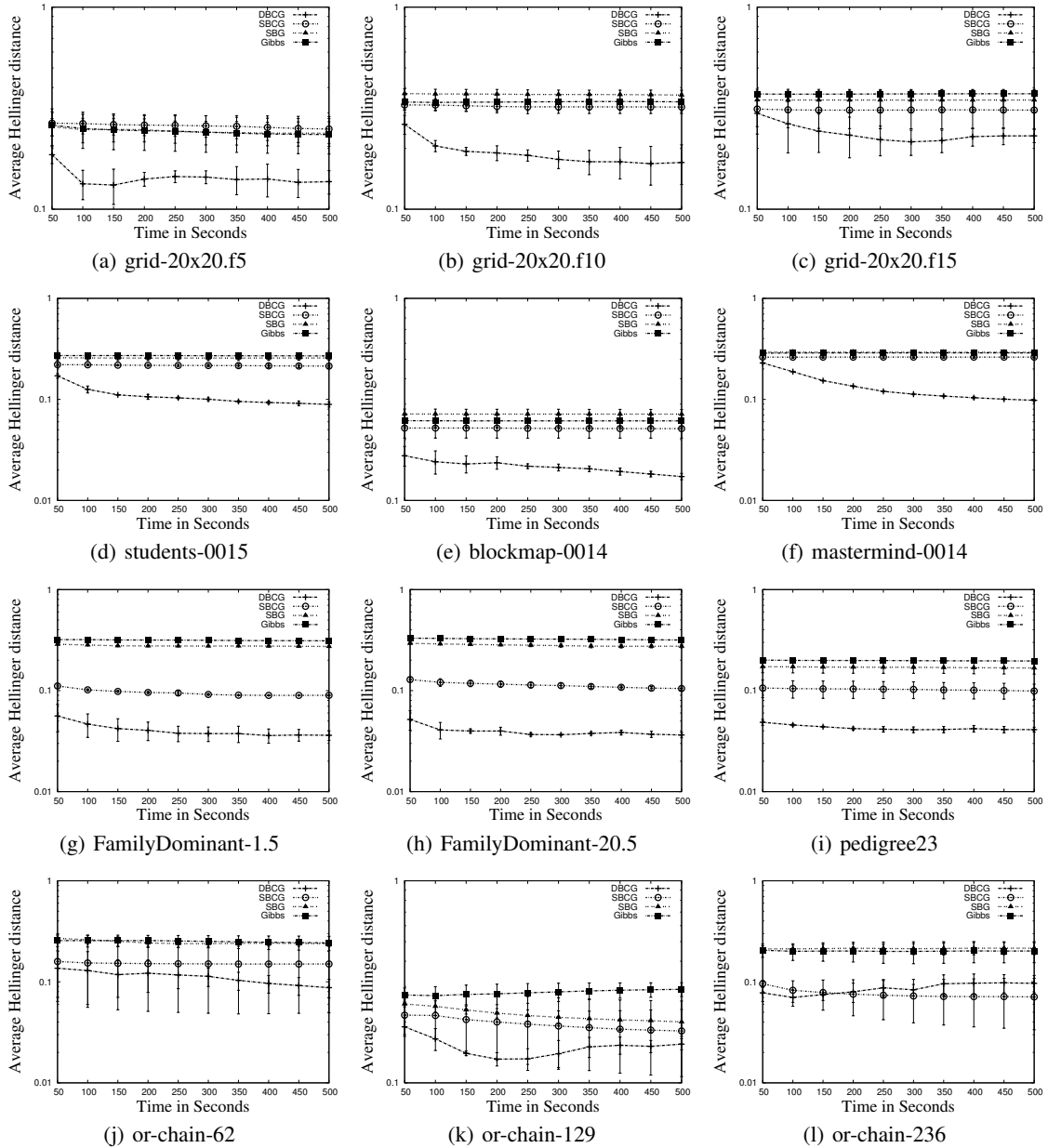

Figure 3: Average Hellinger distance between the exact and the approximate 1-variable marginals plotted as a function of time. (a)-(c): Grids, (d)-(g): Relational, (h)-(i): Linkage, (j)-(l): Promedas.

Fig. 3 shows the results. We see that DBCG is more accurate than all other algorithms on almost all the PGMs, often outperforming the competition by an order of magnitude.

**Ising models.** Figs. 3(a)-(c) show the performance of various algorithms on three Ising models of size 20×20 with evidence on 5, 10 and 15 randomly selected nodes respectively. DBCG is the best algorithm on all three PGMs. SBCG performs better than the other two algorithms on grid20x20.f10 and grid20x20.f15 and its performance is almost similar to SBG and Gibbs on grid20x20.f5.

**Relational** PGMs are formed by grounding statistical relational models [20]. Statistical relational models such as Markov logic networks [21, 22] often have large number of correlated variables as well as deterministic dependencies. Our dynamic approach is beneficial on such models because it has the ability to learn correlations and adjust the partitions accordingly. We experimented with three relational PGMs available from the UAI-08 repository: students-0015, blockmap-0014 and mastermind-0014. Figs. 3 (d)-(f) show the results. Again, we see that DBCG is the best performer followed by SBCG.

**Linkage** PGMs are used for performing genetic linkage analysis [23]. Figs. 3 (g)-(i) show results on three linkage PGMs. Again, on all three PGMs, DBCG is the best

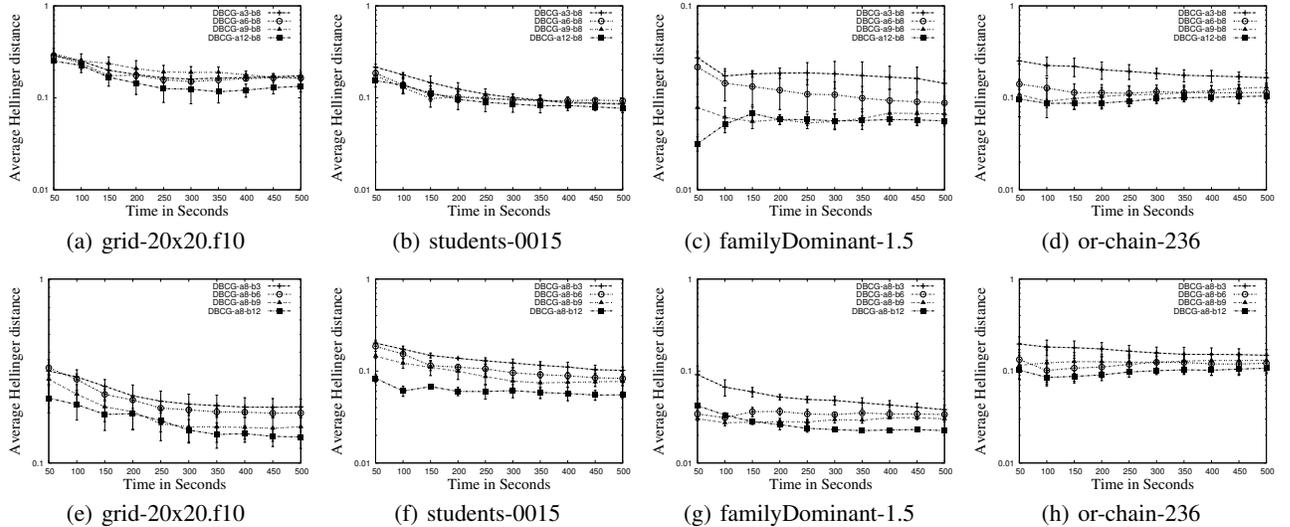

Figure 4: Blocking vs. Collapsing tradeoff. (a)-(d): Impact of varying $\alpha$ with $\beta$ set to a constant value. (e)-(f): Impact of varying $\beta$ with $\alpha$ set to a constant value. We use $\gamma = 50 \times \alpha$. In all the plots, we plot the average Hellinger distance between the exact and the approximate 1-variable marginals as a function of time. The notation shown in the plots is as follows. DBCG-a$x$-b$y$ indicates that $\alpha = x$ and $\beta = y$.

performing algorithm and SBCG is the second best.

**Promedas** PGMs are noisy-OR medical diagnosis networks generated by the Promedas medical diagnosis system [24]. The networks are two-layered bi-partite graphs in which bottom layer has the symptoms and the top layer has the diseases. We experimented with three PGMs: or-chain-62, or-chain-129 and or-chain-236. Figs. 3 (j)-(l) show the results. DBCG performs better than all other algorithms in two out of the three PGMs. On or-chain-236, SBCG is slightly better than DBCG, but has larger variance.

### 6.1 Impact of varying the parameters $\alpha$ and $\beta$

Fig. 4 shows the impact of changing the parameters $\alpha$ and $\beta$ on the performance of DBCG. For brevity, we show results on only one problem instance from each domain. Figs. 4 (a)-(d) show the impact of increasing $\alpha$ with $\beta$ set to a constant while Figs. 4 (e)-(h) show the impact of increasing $\beta$ with $\alpha$ set to a constant. We see that increasing $\alpha$ or $\beta$ typically increases the accuracy and reduces the variance as a function of time. However, in some cases (e.g., Fig. 4(c) and Fig. 4(g)), we see that the accuracy goes down as we increase $\alpha$ and $\beta$, which indicates that there is a tradeoff between blocking and collapsing. In summary, $\alpha$ and $\beta$ help us explore the region between a completely collapsed and a completely blocked sampler, and in turn help us achieve the right balance between blocking and collapsing.

## 7 Summary

In this paper, we formulated the problem of combining blocking and collapsing in computation-limited settings as a multi-objective optimization problem. We proposed a greedy algorithm to solve this problem. The greedy algorithm assumes access to correlations between all pairs of variables. Since the exact value of these correlations is not available, we proposed to estimate them from the generated samples, and update the greedy solution periodically. This yields a dynamic blocked collapsed Gibbs sampling algorithm which iterates between two steps: partitioning and sampling. In the partitioning step, the algorithm uses the current estimate of correlations between variables to partition the variables in the PGM into blocked and collapsed subsets and constructs the collapsed PGM. In the sampling step, the algorithm uses the blocks constructed in the previous step to generate samples from the collapsed PGM and updates the estimate of the 1-variable marginals and the correlations between variables. We performed a preliminary experimental study comparing the performance of our dynamic algorithm with static graph-based blocked collapsed Gibbs sampling algorithms. Our results clearly demonstrated the power and promise of our new approach: in many cases, our dynamic algorithm was an order of magnitude better in terms of accuracy than static graph-based algorithms.

### Acknowledgements

This research was partly funded by the ARO MURI grant W911NF-08-1-0242. The views and conclusions contained in this document are those of the authors and should not be interpreted as representing the official policies, either expressed or implied, of ARO or the U.S. Government.